\documentclass[]{spie}

\usepackage{amsmath,amsfonts,amssymb}
\usepackage{graphicx}
\usepackage{cite}
\usepackage{times}
\usepackage[colorlinks=true,allcolors=blue]{hyperref}

\title{Benchmarking Active Spot Selection for Cost-Efficient Spatial Transcriptomics}

\author[a]{Zheyu Zhu}
\author[b]{Junchao Zhu}
\author[c]{Fengbei Liu}
\author[b]{Tianyuan Yao}
\author[d]{Gelei Xu}
\author[e]{John Cannon}
\author[f]{Haichun Yang}
\author[b]{Yuankai Huo}
\author[c,g]{Mert R. Sabuncu}
\author[g]{Ruining Deng}

\affil[a]{University of Pennsylvania, Philadelphia, PA 19104, USA}
\affil[b]{Vanderbilt University, Nashville, TN, USA}
\affil[c]{Cornell Tech, New York, NY 10044, USA}
\affil[d]{University of Notre Dame, South Bend, IN 46556, USA}
\affil[e]{New York Medical College, Valhalla, NY 10595, USA}
\affil[f]{Vanderbilt University Medical Center, Nashville, TN 37232, USA}
\affil[g]{Weill Medical College of Cornell University, New York, NY 10065, USA}

\authorinfo{Corresponding author: Ruining Deng\\
E-mail: rud4004@med.cornell.edu}

\begin{document}
\maketitle

\begin{abstract}
Spatial transcriptomics (ST) measures gene expression while preserving tissue organization, but dense predefined capture grids can be costly and may repeatedly sample morphologically similar regions. Active learning (AL) may reduce the number of measured spots by selecting an informative subset. However, most AL strategies were developed for categorical labels and independent samples, whereas ST involves high-dimensional continuous expression vectors, spatially correlated candidates, and multiple evaluation measures. We conduct a retrospective pool-based benchmark to determine whether active spot selection improves over uniform Random sampling. Starting from two fully profiled public ST cohorts, we mask candidate expression vectors and simulate multi-round selection using two uncertainty-based strategies, Monte Carlo dropout (MC-dropout) and temporal output discrepancy (TOD), and two diversity-based strategies, CoreSet and TypiClust-inspired selection. The budgeted comparison contains 160 completed configurations across budgets of 5\%, 10\%, 30\%, and 50\% of the fold-wide training spot pool under patient-level cross-validation; a separate full-label reference is also reported. Within each budget, strategies share the selection schedule, morphology-to-expression predictor, and optimization protocol. Performance is assessed using mean per-gene within-slide Pearson correlation coefficient (PCC), expression-cluster agreement, and Moran's~$I$ fidelity. On HER2-positive breast cancer, the pooled mean PCC difference across the four active strategies was negative at 5\% ($-0.0176$) and 10\% ($-0.0117$), and positive but small at 30\% ($+0.0056$) and 50\% ($+0.0057$). On cutaneous squamous cell carcinoma (cSCC), three strategies were below Random at 5\%, and all four were below Random at 10\%. On HER2-positive breast cancer, CoreSet and MC-dropout had lower PCC but higher expression-cluster agreement than Random at the two smallest budgets; this pattern did not reproduce on cSCC. Under the reported fixed training horizons, the evaluated active strategies therefore do not consistently improve on Random sampling at small budgets, and their rankings depend on the evaluation measure.
\end{abstract}

\keywords{spatial transcriptomics, active learning, spot selection, gene-expression prediction, benchmarking}

\section{Introduction}
\label{sec:intro}

\begin{figure}[htbp]
\centering
\includegraphics[width=\textwidth]{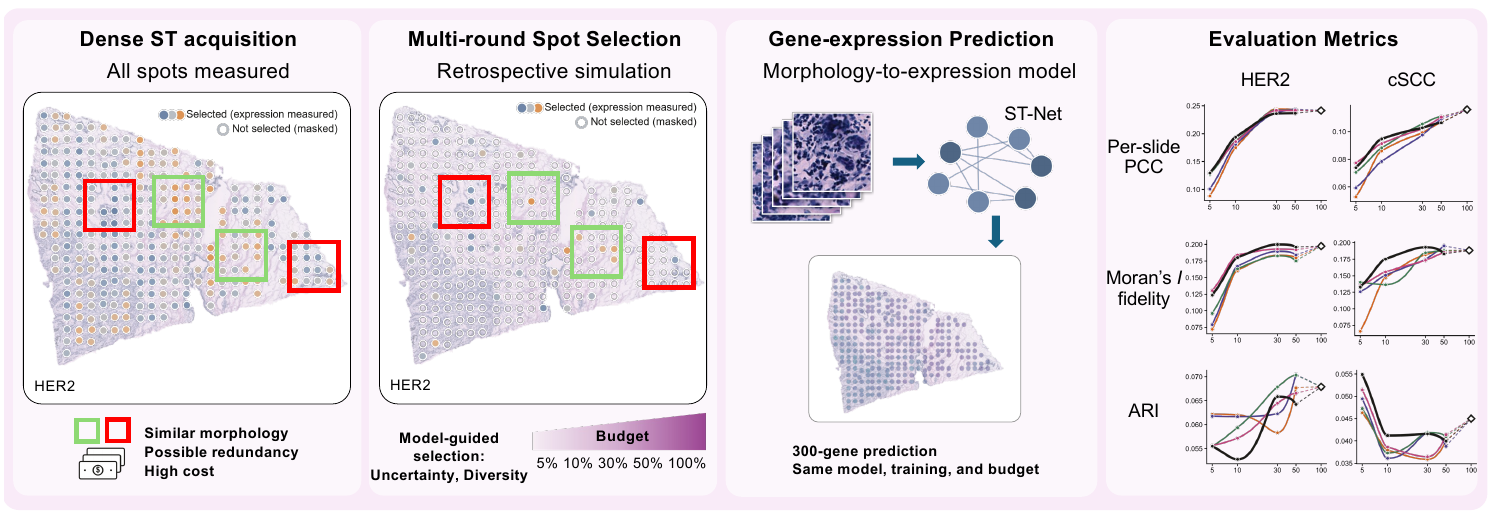}
\caption{Overview of the retrospective benchmark. Expression measurements are hidden in a fully profiled tissue section and revealed only for selected spots. A fixed morphology-to-expression predictor is trained as spots are selected over multiple rounds, and predictions are evaluated on held-out patients. The four budgets are separate experiments with budget-specific training horizons. The outlined regions in the first two panels illustrate morphologically similar tissue and motivate testing, rather than assuming, measurement redundancy. The benchmark holds the predictor, acquisition schedule, spot budget, and optimization protocol fixed across strategies, while residual stochastic training variation remains. The evaluation measures are defined in Section~\ref{sec:evaluation}, and the results are reported in Section~\ref{sec:results}.}
\label{fig:overview}
\end{figure}

Spatial transcriptomics (ST) measures gene expression while preserving the spatial organization of tissue, supporting studies of tumor architecture, immune infiltration, and tissue development~\cite{stahl2016visualization,ji2020multimodal,andersson2021spatial,zhu2026comprehensive}. On commonly used barcoded-array platforms, tissue sections are profiled over a predefined grid of capture spots, including spots in morphologically similar regions. Because ST acquisition is substantially more expensive than routine whole-slide imaging~\cite{zhu2024asign}, dense acquisition can limit the number of patients and sections included in a study. Spatial gene expression can also be autocorrelated, so neighboring spots in histologically homogeneous regions may exhibit similar expression patterns. These properties motivate a practical question: can a selected subset of spots provide sufficient supervision for learning the relationship between tissue morphology and gene expression? Before customized capture regions or region-selective assays are developed, this premise can be tested retrospectively by comparing selected and uniformly sampled spot subsets of the same size.

Previous studies have developed many approaches for predicting spatial gene expression from hematoxylin and eosin (H\&E) images. ST-Net formulated patch-to-spot regression using a pretrained convolutional neural network~\cite{he2020integrating}; later methods incorporated spatial context through transformers and graph neural networks~\cite{pang2021leveraging,zeng2022spatial}, exemplar retrieval and contrastive learning~\cite{yang2023exemplar,xie2023bleep}, pathology foundation models~\cite{chen2024uni}, and other architectures~\cite{zhu2025magnet,zhu2025img2st,zhu2026duet}. SCR$^2$-ST instead considers efficient active sampling through reinforcement learning~\cite{zhu2025scr2}. The present study addresses a narrower benchmarking gap: under matched budgets and a shared predictor, how do standard uncertainty- and diversity-based acquisition strategies compare with Random sampling? This setting differs from conventional active learning (AL) because each spot has a high-dimensional continuous expression vector, neighboring candidates are spatially correlated, and a subset that improves pointwise expression prediction may not preserve spatial expression structure. AL performance can also depend on the available budget~\cite{hacohen2022active}, making evaluation at multiple budgets necessary.

Figure~\ref{fig:overview} summarizes our retrospective pool-based benchmark. Starting from two fully profiled public cohorts, we mask the expression vectors of training candidates and simulate multi-round selection. We compare Random sampling with Monte Carlo dropout (MC-dropout)~\cite{gal2016dropout,gal2017deep}, temporal output discrepancy (TOD)~\cite{huang2021semi}, CoreSet~\cite{sener2018active}, and a TypiClust-inspired strategy~\cite{hacohen2022active} at 5\%, 10\%, 30\%, and 50\% of the fold-wide training spot pool. All strategies use the same patient-level splits, selection schedule, morphology-to-expression predictor, and optimization protocol. We evaluate mean per-gene within-slide Pearson correlation coefficient (PCC), expression-cluster agreement, and Moran's~$I$ fidelity to determine whether the relative performance of active selection depends on the spot budget and evaluation measure.

\section{Methods}
\label{sec:method}

\subsection{Retrospective Active Spot-Selection Benchmark}
\label{sec:benchmark}

We retrospectively simulate spot acquisition by masking expression labels in fully profiled ST cohorts. For a training fold with $N$ candidate spots, spot $i$ has an H\&E patch $x_i$ and a measured expression vector $y_i\in\mathbb{R}^{G}$, with $G=300$. Selecting a spot reveals its complete expression vector, so the budget counts spots rather than genes, reads, or image patches. The expression of unselected spots remains hidden throughout acquisition.

The candidate pool contains all training-fold spots pooled across patients and slides. A strategy can use candidate images and previously revealed expression but cannot access the expression of an unselected spot or data from a held-out patient. Because selection is global rather than slide-specific, the budget is a fraction of the fold-wide training pool and does not enforce equal coverage of each slide.

We evaluate $\beta\in\{0.05,0.10,0.30,0.50\}$ over $R=6$ acquisition rounds in one continuous training run. The first batch is selected by Random sampling because no task-adapted predictor is available. Within a fold and budget, strategies use the same quota, predictor, acquisition schedule, and training schedule. The per-round quota is $q=\lfloor\beta N/R\rfloor$, giving $Rq$ delivered spots. The predictor is not reinitialized between rounds, so later selections depend on the labeled set and preceding optimization. We also report a full-label reference with $\beta=1$. This reference has no acquisition strategy or scoring step and is trained for 50 epochs; it is reported for context rather than as a schedule-matched upper bound.

The benchmark evaluates individual training spots rather than prospective sequencing or contiguous capture regions. The four active strategies use automated scores derived from model predictions or learned image features; they do not incorporate explicit pathologist-defined morphology, tissue-compartment constraints, or clinical knowledge. Consequently, the benchmark tests the computational premise of automated spot selection but not expert-guided acquisition or the feasibility and biological performance of a region-selective assay.

\subsection{Acquisition Strategies}
\label{sec:strategies}

All strategies select without replacement from the fold-wide training pool and use deterministic image preprocessing during candidate scoring.

\noindent\textbf{Random.}
Random samples uniformly without replacement and is the reference baseline. Because sampling occurs over spots rather than slides, the expected number selected from a slide is proportional to its candidate count. Random requires neither model inference nor an acquisition-scoring pass.

\noindent\textbf{MC-dropout.}
MC-dropout~\cite{gal2016dropout,gal2017deep} estimates uncertainty using repeated stochastic forward passes with dropout enabled. Predictive variance is computed for each output gene and aggregated across the $G$ targets; candidates with the largest scores are selected. Batch-normalization layers remain in evaluation mode. Because expression targets are not standardized by gene, gene scale can influence the aggregate acquisition score.

\noindent\textbf{TOD.}
TOD~\cite{huang2021semi} scores candidate $i$ by temporal prediction change:
\begin{equation}
s_{\mathrm{TOD}}(i)=\left\|f_{\theta_t}(x_i)-f_{\theta_{t-1}}(x_i)\right\|_2.
\label{eq:tod}
\end{equation}
As shown in Eq.~\ref{eq:tod}, a large score indicates that a candidate's predicted expression vector changes substantially between two model states. Predictions are cached by spot identifier. When no preceding cache is available, TOD uses Random selection. This implementation adapts the score to 300-output regression.

\noindent\textbf{CoreSet.}
CoreSet~\cite{sener2018active} applies greedy $k$-center selection to learned image features. Previously labeled features are the initial centers, and each step adds the candidate farthest from its nearest center under Euclidean distance. Features are extracted immediately before the regression head and are not $L_2$-normalized. CoreSet therefore measures feature-space coverage, not physical-space coverage.

\noindent\textbf{TypiClust-inspired.}
The TypiClust-inspired strategy~\cite{hacohen2022active} clusters current candidate features and selects a typical member of each cluster using within-cluster neighbor distance. Unlike the published method, this implementation uses task-predictor rather than self-supervised features and sets the cluster count to the round quota instead of $\min(|\mathcal{L}|+B,K_{\max})$. Its first round uses Random sampling because task-adapted features are initially unavailable.

Numerical strategy settings and acquisition timing are reported in Section~\ref{sec:setup}.

\subsection{Shared Morphology-to-Expression Predictor}
\label{sec:predictor}

All strategies use the same ST-Net-style predictor~\cite{he2020integrating}: a DenseNet-121 trunk~\cite{huang2017densely}, global average pooling, a 1,024-unit linear layer, a rectified linear unit, dropout, and a $G$-output regression layer. The trunk is initialized with torchvision's \texttt{IMAGENET1K\_}\allowbreak\texttt{V1} weights. All parameters are trainable, and mean squared error is applied to the targets defined in Section~\ref{sec:setup}. Using one predictor controls the architecture within this benchmark; it does not establish that the acquisition rankings transfer to other backbones or objectives.

\section{Data and Experimental Design}
\label{sec:data}

\subsection{Datasets and Experimental Setup}
\label{sec:setup}

We use two public legacy Spatial Transcriptomics cohorts from HEST-1k~\cite{jaume2024hest}: HER2-positive breast cancer~\cite{andersson2021spatial} and cutaneous squamous cell carcinoma (cSCC)~\cite{ji2020multimodal} (Table~\ref{tab:datasets}).

\begin{table}[htbp]
\centering
\caption{Cohorts used in the conference benchmark. Counts reflect HEST-1k preprocessing.}
\label{tab:datasets}
\resizebox{0.8\textwidth}{!}{%
\begin{tabular}{lccccc}
\hline
Cohort & Patients & Slides & Spots & Retained genes & Targets \\
\hline
HER2-positive breast cancer & 8 & 36 & 13,620 & 967 & 300 \\
cSCC & 4 & 12 & 8,671 & 392 & 300 \\
\hline
\end{tabular}
}
\end{table}

Cross-validation is patient-level. Each of four HER2-positive folds holds out two patients (9 slides; 2,457--3,810 spots), and each cSCC fold holds out one patient (3 slides; 1,632--3,398 spots). Held-out patients are used only for final evaluation.

Each input is a centered $224\times224$-pixel H\&E patch stored at network resolution without resampling. Image resolution is $0.662$--$0.946\,\mu\mathrm{m}$/pixel for HER2-positive breast cancer and $0.545$--$0.554\,\mu\mathrm{m}$/pixel for cSCC. Targets are the first $G=300$ alphabetically ordered genes, with $\tilde{y}_{ig}=\log(1+c_{ig})$ for raw count $c_{ig}$. No library-size normalization, clipping, scaling, or gene-wise standardization is applied.

Five sampling strategies, four budgets, and four folds give 80 completed configurations per cohort and 160 in total; full-label references are separate. Acquisition occurs at epochs 0, 5, 8, 11, 14, and 17 before gradient updates. TOD uses Random at epoch 5 because no preceding cache exists and therefore controls four rounds; the other active strategies control five. With $q=\lfloor\beta N/R\rfloor$, delivered totals can be up to $R-1$ spots below nominal (Table~\ref{tab:protocol}).

\begin{table}[htbp]
\centering
\caption{Budget and training settings for HER2-positive breast cancer fold 1, with $N=9{,}810$ training spots and $R=6$ acquisition rounds.}
\label{tab:protocol}
\begin{tabular}{lccccc}
\hline
$\beta$ & Nominal & Delivered & Per round & Epochs & Optimizer steps \\
\hline
0.05 & 490 & 486 & 81 & 110 & 715 \\
0.10 & 981 & 978 & 163 & 80 & 930 \\
0.30 & 2,943 & 2,940 & 490 & 70 & 2,260 \\
0.50 & 4,905 & 4,902 & 817 & 60 & 3,156 \\
\hline
\end{tabular}
\end{table}

Training uses stochastic gradient descent with momentum 0.9, no Nesterov acceleration, weight decay $10^{-4}$, and batch size 80 with the final partial batch retained. The learning rate is
\begin{equation}
\eta(t)=10^{-4}\left(1-\frac{t}{T_{\max}}\right)^{0.9},
\label{eq:lr}
\end{equation}
where $t$ is the epoch and $T_{\max}$ is the budget-specific horizon. Equation~\ref{eq:lr} runs continuously as the labeled set grows; final-epoch predictions are evaluated. Training applies horizontal and vertical flips independently ($p=0.5$ each), followed by a $90^{\circ}$ rotation ($p=0.5$); scoring is deterministic, without stain normalization or color augmentation. MC-dropout uses five passes with head dropout $p=0.3$ and batch normalization in evaluation mode. TypiClust-inspired uses $k$-means with $n_{\mathrm{init}}=3$ and at most 20 within-cluster neighbors.

All configurations use seed 42 and one NVIDIA RTX 3080 (10~GB) with Python 3.11.15, PyTorch 2.6.0, torchvision 0.21.0, CUDA 12.4, cuDNN 9.1.0, and scikit-learn 1.9.0. Deterministic algorithms are disabled and training randomness is not paired across strategies. Optimizer exposure also increases with budget (715 versus 3,156 steps at 5\% and 50\% in Table~\ref{tab:protocol}); cross-budget comparisons therefore reflect the stated training horizons, not matched optimization or convergence.

\subsection{Evaluation and Reporting}
\label{sec:evaluation}

\noindent\textbf{Gene-expression accuracy.}
Within each held-out slide, PCC is computed across spots per gene and averaged over genes, then aggregated within the fold. Correlations are averaged without Fisher transformation; a $10^{-8}$ denominator stabilizer makes constant genes contribute approximately zero.

\noindent\textbf{Expression-cluster agreement.}
Predicted and measured profiles are clustered independently within each slide using $k$-means ($k=10$, $n_{\mathrm{init}}=10$, seed 42) without scaling or dimensionality reduction. Adjusted Rand index (ARI)~\cite{hubert1985comparing} compares the partitions; slides with fewer than $2k$ spots are excluded. This measures expression-cluster agreement, not histological domain recovery.

\noindent\textbf{Moran's $I$ fidelity.}
Moran's $I$~\cite{moran1950notes} is computed per slide and gene using a symmetrized six-nearest-neighbor graph on array coordinates with binary, non-row-standardized weights. Fidelity is the PCC across genes between measured and predicted Moran's $I$; at least three defined genes are required.

\noindent\textbf{Statistical reporting.}
We report active-minus-Random differences and fold sign counts descriptively. Because folds overlap, strategies share the Random comparator, and each configuration has one seed, pooled $p$-values are not reported.

\noindent\textbf{Acquisition cost.}
Acquisition cost is per-round candidate-scoring time, including required feature extraction but excluding training. Random has no scoring pass; timings are implementation-specific.

\section{Results}
\label{sec:results}

\subsection{Gene-Expression Accuracy Across Budgets}

PCC increases with budget on both cohorts (Tables~\ref{tab:pcc_her2} and~\ref{tab:pcc_cscc}), but fold variation exceeds the mean differences among strategies. The 5\% and 10\% models are still improving at their final epochs, whereas changes are below 2.5\% at 30\% and 50\%; comparisons therefore apply to the reported training horizons. The full-label reference reaches $0.2416\pm0.1460$ on HER2-positive breast cancer and $0.1160\pm0.0220$ on cSCC. Random at 5\% reaches 54\% and 64\% of these means, and Random at 50\% reaches 98\% and 92\%, respectively. These ratios are descriptive because the 50-epoch full-label reference is not schedule-matched to the 60-epoch 50\% runs.

\begin{table}[htbp]
\centering
\caption{HER2-positive breast cancer PCC, mean $\pm$ SD over four folds. Bold marks the largest mean per budget; cost is the range of fold-median scoring times at 50\%.}
\label{tab:pcc_her2}
\resizebox{\textwidth}{!}{%
\begin{tabular}{lccccr}
\hline
Strategy & 5\% & 10\% & 30\% & 50\% & Cost/round \\
\hline
Random & $0.1295 \pm 0.0680$ & $\mathbf{0.1940} \pm 0.1040$ & $0.2361 \pm 0.1340$ & $0.2367 \pm 0.1380$ & $<1$\,s \\
CoreSet~\cite{sener2018active} & $0.0883 \pm 0.0500$ & $0.1751 \pm 0.1000$ & $\mathbf{0.2442} \pm 0.1300$ & $0.2422 \pm 0.1400$ & $11$\,s \\
MC-dropout~\cite{gal2016dropout} & $0.1009 \pm 0.0580$ & $0.1811 \pm 0.1010$ & $0.2378 \pm 0.1300$ & $\mathbf{0.2439} \pm 0.1400$ & $38$\,s \\
TOD~\cite{huang2021semi} & $0.1276 \pm 0.0870$ & $0.1870 \pm 0.1110$ & $0.2428 \pm 0.1360$ & $0.2418 \pm 0.1430$ & $8$\,s \\
TypiClust-inspired~\cite{hacohen2022active} & $\mathbf{0.1308} \pm 0.0700$ & $0.1857 \pm 0.0990$ & $0.2418 \pm 0.1350$ & $0.2417 \pm 0.1450$ & $71$--$1{,}584$\,s \\
\hline
\multicolumn{6}{l}{\emph{Full-label reference (}$\beta=1$\emph{, no selection, 50 epochs): $0.2416 \pm 0.1460$}} \\
\hline
\end{tabular}
}
\end{table}

\begin{figure}[htbp]
\centering
\includegraphics[width=\textwidth]{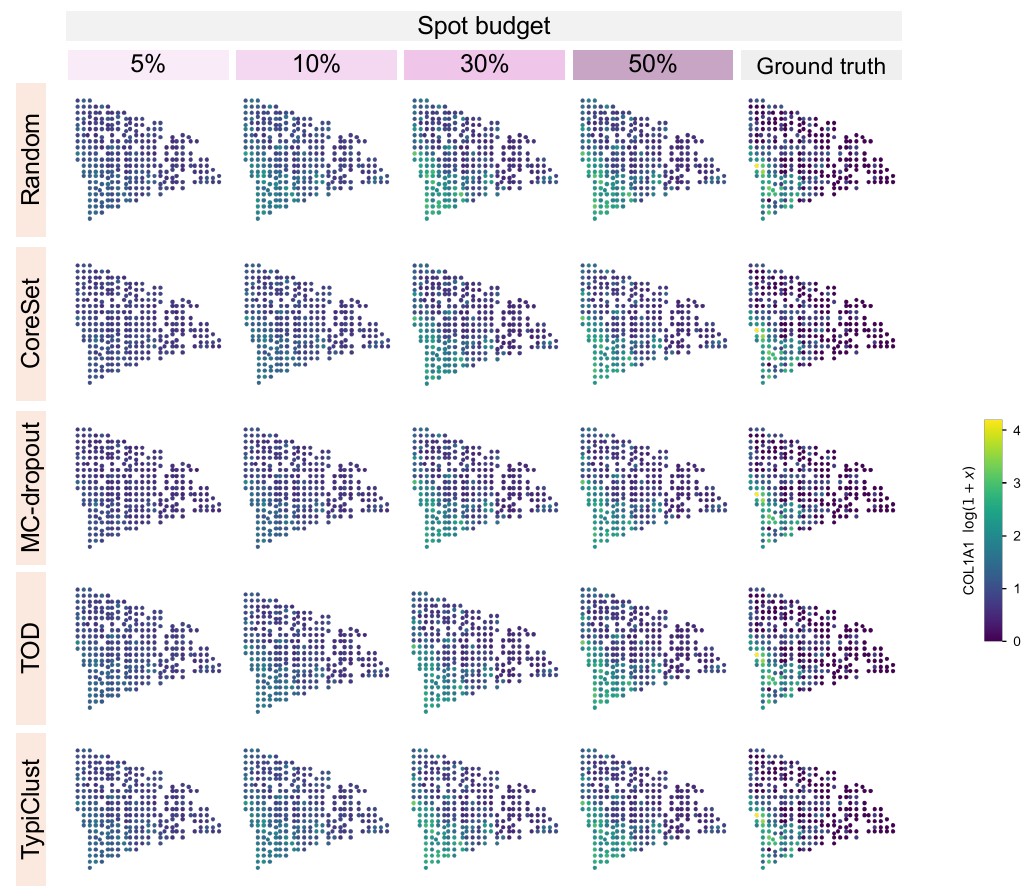}
\caption{Predicted \textit{COL1A1} on held-out HER2-positive slide SPA143. Rows are strategies, columns are budgets, and all panels share the measured-map scale. Predicted SD increases from 0.18--0.28 at 5\% to 0.67--0.70 at 50\%, versus 0.87 measured. This single-slide example is not used to rank strategies.}
\label{fig:predmaps}
\end{figure}

\subsection{Paired Comparison With Random}

On HER2-positive breast cancer, three active strategies are below Random at 5\% and all four at 10\%; all four mean differences are positive but small at 30\% and 50\% (Table~\ref{tab:delta_her2}). The pooled descriptive differences are $-0.0176$, $-0.0117$, $+0.0056$, and $+0.0057$. On cSCC, three strategies are below Random at 5\%, all four at 10\%, and all four slightly exceed Random at 50\% (Table~\ref{tab:pcc_cscc}).

\begin{table}[htbp]
\centering
\caption{HER2-positive paired PCC difference from Random; parentheses give positive folds out of four. Bold marks the largest mean per budget. Pooled rows are descriptive.}
\label{tab:delta_her2}
\resizebox{0.8\textwidth}{!}{%
\begin{tabular}{lcccc}
\hline
Strategy & 5\% & 10\% & 30\% & 50\% \\
\hline
CoreSet~\cite{sener2018active} & $-0.0412$ (0/4) & $-0.0189$ (1/4) & $\mathbf{+0.0081}$ (2/4) & $+0.0055$ (4/4) \\
MC-dropout~\cite{gal2016dropout} & $-0.0286$ (1/4) & $-0.0129$ (1/4) & $+0.0017$ (1/4) & $\mathbf{+0.0072}$ (3/4) \\
TOD~\cite{huang2021semi} & $-0.0019$ (2/4) & $\mathbf{-0.0069}$ (2/4) & $+0.0067$ (3/4) & $+0.0051$ (3/4) \\
TypiClust-inspired~\cite{hacohen2022active} & $\mathbf{+0.0013}$ (2/4) & $-0.0082$ (0/4) & $+0.0057$ (4/4) & $+0.0050$ (2/4) \\
\hline
\multicolumn{5}{l}{\emph{Pooled descriptively over all four active strategies ($n=16$ strategy-fold pairs per budget)}} \\
Mean $\Delta$ & $-0.0176$ & $-0.0117$ & $+0.0056$ & $+0.0057$ \\
Active strategy better & 5/16 & 4/16 & 10/16 & 12/16 \\
\hline
\end{tabular}
}
\end{table}

\begin{table}[htbp]
\centering
\caption{cSCC PCC, mean $\pm$ SD over four folds. Bold marks the largest mean per budget.}
\label{tab:pcc_cscc}
\resizebox{0.95\textwidth}{!}{%
\begin{tabular}{lcccc}
\hline
Strategy & 5\% & 10\% & 30\% & 50\% \\
\hline
Random & $0.0740 \pm 0.0253$ & $\mathbf{0.0948} \pm 0.0233$ & $0.1029 \pm 0.0212$ & $0.1066 \pm 0.0209$ \\
CoreSet~\cite{sener2018active} & $0.0528 \pm 0.0260$ & $0.0862 \pm 0.0233$ & $0.0994 \pm 0.0130$ & $0.1068 \pm 0.0169$ \\
MC-dropout~\cite{gal2016dropout} & $0.0594 \pm 0.0266$ & $0.0785 \pm 0.0192$ & $0.0977 \pm 0.0181$ & $0.1093 \pm 0.0228$ \\
TOD~\cite{huang2021semi} & $0.0705 \pm 0.0193$ & $0.0883 \pm 0.0173$ & $\mathbf{0.1056} \pm 0.0215$ & $\mathbf{0.1109} \pm 0.0214$ \\
TypiClust-inspired~\cite{hacohen2022active} & $\mathbf{0.0774} \pm 0.0270$ & $0.0915 \pm 0.0230$ & $0.1024 \pm 0.0229$ & $0.1106 \pm 0.0215$ \\
\hline
\multicolumn{5}{l}{\emph{Full-label reference (}$\beta=1$\emph{, no selection, 50 epochs): $0.1160 \pm 0.0220$}} \\
\hline
\end{tabular}
}
\end{table}

\subsection{Spatial-Expression Structure}

At 5\% and 10\% on HER2-positive breast cancer, CoreSet and MC-dropout have lower PCC but higher ARI than Random on all four folds (Table~\ref{tab:spatial}); Figure~\ref{fig:predmaps} illustrates one low-budget prediction. This reversal does not reproduce on cSCC, where every active strategy has lower ARI than Random at both budgets (Table~\ref{tab:spatial_cscc}). Moran's $I$ fidelity also has no consistently best active strategy: Random leads at all HER2-positive budgets except the two small TypiClust-inspired mean differences, and the cSCC leader varies with budget.

\begin{table}[htbp]
\centering
\caption{HER2-positive spatial measures over four folds. Parentheses show difference from Random and positive folds; bold marks the largest mean per metric and budget.}
\label{tab:spatial}
\resizebox{\textwidth}{!}{%
\begin{tabular}{lcccc}
\hline
Strategy & 5\% & 10\% & 30\% & 50\% \\
\hline
\multicolumn{5}{l}{\emph{Expression-cluster agreement (ARI)}} \\
Random & $0.0555$ & $0.0528$ & $0.0658$ & $0.0642$ \\
CoreSet~\cite{sener2018active} & $\mathbf{0.0622}$ ($+0.0067$, 4/4) & $\mathbf{0.0620}$ ($+0.0093$, 4/4) & $0.0583$ ($-0.0075$, 1/4) & $0.0677$ ($+0.0035$, 2/4) \\
MC-dropout~\cite{gal2016dropout} & $0.0618$ ($+0.0063$, 4/4) & $0.0617$ ($+0.0089$, 4/4) & $0.0623$ ($-0.0035$, 1/4) & $\mathbf{0.0705}$ ($+0.0062$, 2/4) \\
TOD~\cite{huang2021semi} & $0.0555$ ($+0.0000$, 1/4) & $0.0594$ ($+0.0066$, 4/4) & $\mathbf{0.0679}$ ($+0.0020$, 3/4) & $0.0704$ ($+0.0061$, 2/4) \\
TypiClust-inspired~\cite{hacohen2022active} & $0.0557$ ($+0.0002$, 2/4) & $0.0572$ ($+0.0044$, 2/4) & $0.0645$ ($-0.0013$, 1/4) & $0.0665$ ($+0.0023$, 1/4) \\
\hline
\multicolumn{5}{l}{\emph{Moran's $I$ fidelity}} \\
Random & $\mathbf{0.1237}$ & $0.1802$ & $\mathbf{0.1999}$ & $\mathbf{0.1962}$ \\
CoreSet~\cite{sener2018active} & $0.0720$ ($-0.0516$, 1/4) & $0.1611$ ($-0.0191$, 0/4) & $0.1836$ ($-0.0163$, 1/4) & $0.1800$ ($-0.0162$, 0/4) \\
MC-dropout~\cite{gal2016dropout} & $0.0794$ ($-0.0443$, 1/4) & $0.1671$ ($-0.0131$, 0/4) & $0.1896$ ($-0.0103$, 1/4) & $0.1843$ ($-0.0119$, 0/4) \\
TOD~\cite{huang2021semi} & $0.0962$ ($-0.0274$, 1/4) & $0.1635$ ($-0.0168$, 1/4) & $0.1826$ ($-0.0173$, 1/4) & $0.1752$ ($-0.0210$, 0/4) \\
TypiClust-inspired~\cite{hacohen2022active} & $0.1304$ ($+0.0068$, 1/4) & $\mathbf{0.1843}$ ($+0.0041$, 2/4) & $0.1929$ ($-0.0071$, 1/4) & $0.1919$ ($-0.0044$, 2/4) \\
\hline
\multicolumn{5}{l}{\emph{Full-label reference (}$\beta=1$\emph{, no selection, 50 epochs): ARI $0.0679$; Moran's $I$ fidelity $0.1975$}} \\
\hline
\end{tabular}
}
\end{table}

\begin{table}[htbp]
\centering
\caption{cSCC spatial measures, reported as in Table~\ref{tab:spatial}. Bold marks the largest mean per metric and budget.}
\label{tab:spatial_cscc}
\resizebox{\textwidth}{!}{%
\begin{tabular}{lcccc}
\hline
Strategy & 5\% & 10\% & 30\% & 50\% \\
\hline
\multicolumn{5}{l}{\emph{Expression-cluster agreement (ARI)}} \\
Random & $\mathbf{0.0549}$ & $\mathbf{0.0412}$ & $0.0416$ & $0.0400$ \\
CoreSet~\cite{sener2018active} & $0.0464$ ($-0.0086$, 1/4) & $0.0380$ ($-0.0032$, 2/4) & $0.0359$ ($-0.0057$, 0/4) & $0.0401$ ($+0.0001$, 3/4) \\
MC-dropout~\cite{gal2016dropout} & $0.0495$ ($-0.0054$, 1/4) & $0.0361$ ($-0.0051$, 0/4) & $0.0417$ ($+0.0001$, 2/4) & $0.0388$ ($-0.0012$, 2/4) \\
TOD~\cite{huang2021semi} & $0.0473$ ($-0.0076$, 0/4) & $0.0373$ ($-0.0039$, 1/4) & $\mathbf{0.0419}$ ($+0.0003$, 2/4) & $0.0410$ ($+0.0010$, 2/4) \\
TypiClust-inspired~\cite{hacohen2022active} & $0.0515$ ($-0.0035$, 1/4) & $0.0387$ ($-0.0026$, 1/4) & $0.0365$ ($-0.0051$, 2/4) & $\mathbf{0.0414}$ ($+0.0015$, 3/4) \\
\hline
\multicolumn{5}{l}{\emph{Moran's $I$ fidelity}} \\
Random & $0.1330$ & $\mathbf{0.1754}$ & $\mathbf{0.1931}$ & $0.1831$ \\
CoreSet~\cite{sener2018active} & $0.0660$ ($-0.0670$, 1/4) & $0.1473$ ($-0.0281$, 1/4) & $0.1817$ ($-0.0114$, 0/4) & $0.1882$ ($+0.0051$, 3/4) \\
MC-dropout~\cite{gal2016dropout} & $0.1257$ ($-0.0073$, 1/4) & $0.1508$ ($-0.0246$, 1/4) & $0.1749$ ($-0.0181$, 0/4) & $\mathbf{0.1952}$ ($+0.0120$, 4/4) \\
TOD~\cite{huang2021semi} & $\mathbf{0.1404}$ ($+0.0074$, 1/4) & $0.1365$ ($-0.0388$, 0/4) & $0.1850$ ($-0.0081$, 1/4) & $0.1877$ ($+0.0046$, 3/4) \\
TypiClust-inspired~\cite{hacohen2022active} & $0.1349$ ($+0.0019$, 3/4) & $0.1558$ ($-0.0196$, 1/4) & $0.1735$ ($-0.0196$, 0/4) & $0.1853$ ($+0.0022$, 3/4) \\
\hline
\multicolumn{5}{l}{\emph{Full-label reference (}$\beta=1$\emph{, no selection, 50 epochs): ARI $0.0450$; Moran's $I$ fidelity $0.1883$}} \\
\hline
\end{tabular}
}
\end{table}

\section{Discussion}
\section{Discussion and Future Work}
\label{sec:discussion}

Under the reported horizons, most active strategies trail Random at 5\% and 10\%, while mean PCC gains at larger budgets are small. Selection of uncommon candidates may contribute, but selected regions were not analyzed and unequal optimization exposure is an alternative explanation. The HER2-positive PCC/ARI reversal also fails to reproduce on cSCC, supporting separate reporting of pointwise and spatial measures.

The comparisons are descriptive because four overlapping folds and one seed do not separate acquisition from training or sampling variation. Small-budget models are not converged, targets are unnormalized $\log(1+c_{ig})$ counts, and the full-label reference is not schedule-matched. Results cover two cohorts on one platform, one ST-Net-style DenseNet-121 backbone, and four automated model-based strategies. Future work should test additional cohorts, tissue types, and ST platforms; stronger predictors, including pathology foundation models; and acquisition rules incorporating explicit morphological constraints, pathologist input, or clinical knowledge. These extensions may change the strategy rankings, but the present results do not establish that they will improve performance. Prospective and contiguous-region acquisition also remain untested, so the observed crossover is not a general budget threshold.

\section{New or Breakthrough Work to Be Presented}

This work provides a controlled retrospective comparison of four standard active-selection strategies and Random sampling for ST spot acquisition. The benchmark evaluates matched spot budgets under one shared predictor and reports both pointwise and spatial-expression measures. The results show that most active strategies underperform Random at small budgets under the reported training horizons, while the small advantages at larger budgets depend on cohort and evaluation measure.

\section{Conclusion}

We evaluated whether four active strategies select more useful ST training spots than Random sampling under matched spot budgets, acquisition schedules, predictors, and optimization protocols. At 5\% and 10\% of the fold-wide training pool, most active-strategy PCC means are below Random; at 30\% and 50\%, the mean differences are positive but small relative to fold-to-fold variation. On HER2-positive breast cancer, the low-budget ranking differs between PCC and expression-cluster agreement, but this reversal does not reproduce on cSCC. The separate full-label reference provides context for the budgeted results but is not schedule-matched. Overall, the present retrospective evidence does not show a consistent advantage for the evaluated active strategies at small budgets. Future region-selective ST studies should report the acquisition budget, training horizon, and evaluation measure and should validate computational selection results in a physically realizable assay.

\section{ACKNOWLEDGMENTS}
This research was supported by the WCM Radiology AIMI Fellowship and WCM CTSC 2027 Pilot Award. Additional support was provided by NIH 1R01HL174863-01A1 (Sabuncu) and NIH 1U54DK144866-01 (Sabuncu).

\bibliography{main}
\bibliographystyle{spiebib}

\end{document}